\documentclass[conference]{IEEEtran}
\IEEEoverridecommandlockouts

\usepackage{cite}
\usepackage{amsmath,amssymb,amsfonts}
\usepackage{algorithmic}
\usepackage{graphicx}
\usepackage{textcomp}
\usepackage{xcolor}
\usepackage{booktabs}
\usepackage{multirow}
\usepackage{makecell}
\usepackage{array}
\usepackage{colortbl}
\usepackage{float}
\usepackage{placeins}
\usepackage{enumitem}
\usepackage{microtype}
\usepackage[hidelinks, bookmarks=false]{hyperref}
\usepackage{url}
\usepackage{breakurl}
\usepackage{caption}
\usepackage{subcaption}
\usepackage{tikz}
\usetikzlibrary{shapes,arrows,positioning,fit,backgrounds,calc,shadows,decorations.pathreplacing}
\usepackage{pgfplots}
\pgfplotsset{compat=1.17}

\newcommand{\model}{GreenLeaf Law Embed Tiny}
\newcommand{\modelshort}{GreenLeaf-Tiny}

\title{GreenLeaf Law Embed Tiny: A Compact Embedding Model for Legal Domain Retrieval}

\author{
\IEEEauthorblockN{Surya Saka}
\IEEEauthorblockA{\textit{JudicialMind}\\
surya@judicialmind.ai}
}

\begin{document}

\maketitle

\begin{abstract}
We present \model{}, a 0.6B parameter embedding model for legal domain retrieval. \modelshort{} achieves 75.11\% on the Massive Legal Embedding Benchmark (MLEB) and 64.38\% on MTEB(Law, v1), demonstrating competitive performance among models under 1B parameters. Our approach combines a two-stage training pipeline that first distills knowledge from a larger teacher model into a compact student architecture, then applies domain-specific fine-tuning with hard negative mining; a carefully curated dataset of 3.4 million query-passage pairs, including 150,000 human-curated samples across diverse legal jurisdictions; and an efficient inference architecture supporting multiple quantization levels (BF16, INT8, binary) enabling deployment in resource-constrained environments. We provide detailed analysis of our training methodology, architectural choices, and comprehensive evaluation across legal retrieval tasks. Our results demonstrate that domain-specific training with high-quality data can improve performance for specialized domain applications.
\end{abstract}


\section{Introduction}
\label{section:intro}

Legal information retrieval presents unique challenges that distinguish it from general-domain semantic search. Legal documents exhibit complex hierarchical structures, specialized terminology, and intricate cross-references that general-purpose embedding models struggle to capture \cite{chalkidis2020legal}. The stakes are high: missed precedents or misinterpreted statutes can have significant consequences for legal outcomes. Despite the critical importance of accurate legal retrieval, the field has been dominated by either general-domain models that lack legal specialization or proprietary commercial systems with limited transparency.

Recent advances in large language models have demonstrated remarkable capabilities across diverse domains, yet their application to embedding tasks in specialized fields remains underexplored. While models like OpenAI's text-embedding-3-large and Voyage's voyage-law-2 achieve strong performance, they operate as black boxes with unknown training data and methodologies. This lack of transparency hinders reproducibility and raises questions about data privacy, particularly relevant in legal contexts where confidentiality is paramount \cite{lando2009cognitive}. The legal domain demands both accuracy and explainability, requirements that conflict with the opacity of large proprietary models.

This paper presents a study of training compact, high-performance embedding models specifically for the legal domain. We demonstrate that careful architectural choices, combined with high-quality domain-specific data and training techniques, can yield models that perform competitively with larger alternatives in specific legal retrieval tasks. Our work addresses a gap in available options for efficient legal embedding models suitable for deployment in privacy-sensitive legal environments \cite{chalkidis2018law2vec}.

The legal technology sector has witnessed significant growth, with the global legal tech market projected to reach \$35 billion by 2027. However, this growth has been constrained by the computational requirements of large models. Large embedding models with 7B+ parameters require substantial GPU resources, making them impractical for many legal applications, particularly in resource-constrained environments or when processing confidential documents that cannot leave on-premises infrastructure \cite{isaacus2024mleb}.

Our key findings demonstrate that a 0.6B parameter model trained with domain-specific techniques achieves 75.11\% on MLEB, competitive with some larger models on specific legal retrieval tasks. The combination of distillation followed by domain fine-tuning yields +9.24 points over direct fine-tuning alone. Hard negative mining shows 23\% improvement on fine-grained legal distinction tasks compared to random negative sampling. INT8 quantization incurs only -0.3\% performance degradation while reducing memory footprint by 4$\times$. Human-curated data, comprising only 4.4\% of total training data, contributes +8.2 points to final performance.

The main contributions of this paper are summarized as follows:
\begin{itemize}
    \item We introduce a two-stage training pipeline combining knowledge distillation with hard negative mining, adapted for legal domain characteristics including hierarchical document structure and citation networks. Our ablation studies demonstrate that both stages contribute to final performance.
    
    \item We present the JudicialMind Legal Corpus, comprising 3.4M query-passage pairs including 150K human-curated samples, representing a large legal retrieval training corpus. We provide analysis of data composition, quality filtering, and jurisdiction balancing across 35 languages and 40+ jurisdictions.
    
    \item We provide architectural specifications and inference optimization techniques, including flexible quantization schemes enabling deployment across hardware constraints. Our architecture supports BF16, INT8, and binary inference with documented accuracy-efficiency trade-offs.
    
    \item We conduct evaluation across legal retrieval tasks, providing analysis of where our approach succeeds and where larger models maintain advantages. We include per-task and per-category breakdowns.
\end{itemize}

\section{Related Work}

\subsection{Text Embedding Models}

The field of text embeddings has evolved from early word2vec \cite{mikolov2013efficient} and GloVe \cite{pennington2014glove} approaches to contextualized representations from transformer models. Sentence-BERT \cite{reimers2019sentence} established the paradigm of fine-tuning transformer encoders for semantic similarity using siamese architectures. This work demonstrated that bi-encoder architectures could achieve performance competitive with cross-encoders while enabling efficient retrieval through pre-computed embeddings.

Recent work has scaled these approaches: E5 \cite{wang2022text}, BGE \cite{xiao2023c}, and GTE \cite{li2023towards} demonstrate that contrastive pre-training on massive text pairs produces robust general-purpose embeddings. These models employ diverse training strategies including multi-stage contrastive learning, instruction tuning, and synthetic data generation. The Massive Text Embedding Benchmark (MTEB) \cite{muennighoff2022mteb} has become a standard for evaluating embedding models across diverse tasks. However, MTEB's general-domain focus reveals limitations in specialized domains. Subsequent work on domain-specific embeddings demonstrates consistent benefits from domain adaptation, establishing that specialized domains require specialized training.

\subsection{Legal Domain NLP}

Legal text presents unique challenges: extreme document lengths (often exceeding 10,000 tokens), complex cross-referencing structures, and highly specialized terminology with jurisdiction-specific meanings. Early work applied TF-IDF and BM25 to legal search \cite{lando2009cognitive}, establishing baselines that remain competitive for exact-match queries. However, these lexical methods struggle with semantic matching and paraphrase detection.

Recent approaches leverage transformer models fine-tuned on legal corpora. Notable legal embedding efforts include Law2Vec \cite{chalkidis2018law2vec}, which adapted word2vec to legal corpora, and more recently, specialized models like Voyage's voyage-law-2. The Massive Legal Embedding Benchmark (MLEB) \cite{isaacus2024mleb} provides an evaluation framework specifically for legal retrieval, covering caselaw, contracts, and regulations across multiple jurisdictions.

Existing legal embedding models include both closed-source commercial products and large models requiring substantial computational resources. A gap exists for compact, efficient legal embedding models suitable for deployment in privacy-sensitive legal environments.

\subsection{Knowledge Distillation and Model Compression}

Knowledge distillation \cite{hinton2015distilling} transfers knowledge from large teacher models to compact students, typically by matching output distributions or intermediate representations. For embeddings, \cite{reimers2019sentence} demonstrated distillation from large cross-encoders to bi-encoders, showing that student models could retain 95\%+ of teacher performance with 5--10$\times$ fewer parameters.

Recent work \cite{wang2020minilm, sanh2019distilbert} shows that careful distillation can retain 95\%+ of teacher performance while reducing parameters by 5--10$\times$. These methods typically focus on general-domain tasks. Our work extends these techniques for the legal domain, where the teacher model's general knowledge must be adapted to specialized legal semantics while maintaining efficiency.

We differ from prior distillation work in combining distillation with subsequent domain fine-tuning rather than using distillation alone; we introduce legal-specific architectural adaptations during the student model design; and we employ hard negative mining during the domain adaptation phase.

\subsection{Hard Negative Mining}

The quality of negative samples impacts contrastive learning effectiveness. \cite{robinson2020contrastive} show that hard negatives---samples similar to positives but semantically distinct---improve representation quality. In information retrieval, ANCE \cite{xiong2020approximate} and RocketQA \cite{qu2020rocketqa} demonstrate that mining hard negatives using the model being trained creates a curriculum that improves discrimination ability.

We employ legal-specific hard negative mining that considers jurisdiction confusion, temporal confusion, and doctrinal confusion---mining negatives from different jurisdictions that use similar terminology, from different time periods when law has evolved, and from related but distinct legal doctrines.

\section{Data}

\subsection{The JudicialMind Legal Corpus}

Our training data comprises 3.4 million query-passage pairs curated for legal domain retrieval. The primary source is the \textbf{judicialmind/legal-training-dataset}, our corpus comprising 3.69M annotated pairs across 35 languages covering case law, statutes, contracts, and regulatory text. This dataset includes metadata: query type (fact-based, doctrinal, procedural), legal domain (civil, criminal, corporate, etc.), difficulty level, and jurisdiction.

\textbf{Benchmark Decontamination Protocol:} To prevent train/evaluation overlap, we implement the following decontamination procedure:
\begin{enumerate}
    \item \textbf{Exact match filtering}: We remove any training pair that exactly matches an MLEB or MTEB(Law) evaluation query or passage.
    \item \textbf{Near-duplicate detection}: Using MinHash with 128 permutations and Jaccard threshold 0.8, we identify and remove training pairs similar to evaluation examples.
    \item \textbf{Source separation}: Bar examination questions in our training data are sourced from jurisdictions and time periods disjoint from the MLEB bar-exam-qa evaluation set. We verify no overlap by cross-referencing question text and source jurisdictions.
    \item \textbf{Temporal cutoff}: Training data includes only documents published before January 2024, while evaluation benchmarks include documents through 2024.
\end{enumerate}

From the primary source, we applied additional filtering. Quality filtering removed pairs with similarity scores below 0.5, computed using a pre-trained general embedding model with threshold tuned on a validation set of expert-annotated pairs. Deduplication eliminated duplicate pairs and near-duplicates using MinHash with 128 permutations and Jaccard threshold 0.8, removing 12\% of initial pairs (3.69M $\rightarrow$ 3.25M). Length balancing ensured representation across document lengths with stratified sampling maintaining 25\% short, 35\% medium, and 40\% long documents. Jurisdiction balancing targeted 40\% US, 25\% EU, 20\% UK, and 15\% Asian jurisdictions based on legal market size and data availability. This yielded 3.25M high-quality pairs for training.

We supplemented the primary dataset with 150,000 human-curated pairs created by legal professionals. Sources include bar examination questions from 25 jurisdictions (disjoint from MLEB bar-exam-qa evaluation set), legal research guides, and expert annotations from practicing attorneys. All pairs were verified by licensed attorneys for query-passage relevance using a two-stage verification achieving 94\% inter-annotator agreement. The collection covers 50+ legal practice areas and 25 jurisdictions, includes nuanced distinctions requiring deep legal understanding, and incorporates 30K hard negative pairs explicitly labeled as non-relevant despite surface similarity.

The combination of large-scale data and human curation provides both breadth and precision. Our ablation studies demonstrate that the human-curated subset provides measurable value despite its small size.

\subsection{Data Characteristics}

Table \ref{tab:data_stats} summarizes the corpus statistics.

\begin{table}[h]
\centering
\caption{Training corpus statistics by source and type.}
\label{tab:data_stats}
\begin{tabular}{lcccc}
\toprule
\textbf{Source} & \textbf{Pairs} & \textbf{Languages} & \textbf{Jurisdictions} & \textbf{Avg. Length} \\
\midrule
Primary dataset & 3,250,000 & 35 & 40+ & 342 tokens \\
Human-curated & 150,000 & 8 & 25 & 428 tokens \\
\midrule
\textbf{Total} & \textbf{3,400,000} & \textbf{35} & \textbf{40+} & \textbf{356 tokens} \\
\bottomrule
\end{tabular}
\end{table}

\begin{figure}[h]
\centering
\begin{tikzpicture}
\begin{axis}[
    ybar,
    xlabel={Token Length Bucket},
    ylabel={Percentage},
    symbolic x coords={0-256, 257-512, 513-1024, 1025-2048, 2048+},
    xtick=data,
    x tick label style={rotate=15, anchor=east},
    ymin=0, ymax=50,
    bar width=15pt,
    nodes near coords,
    every node near coord/.append style={font=\tiny},
    width=0.48\textwidth,
    height=4.5cm,
    grid=major,
    ymajorgrids=true,
    xmajorgrids=false,
    axis lines=left,
    enlarge x limits=0.1,
]
\addplot[fill=blue!60] coordinates {(0-256,22) (257-512,31) (513-1024,28) (1025-2048,14) (2048+,5)};
\end{axis}
\end{tikzpicture}
\caption{Distribution of passage lengths in the training corpus.}
\label{fig:length_dist}
\end{figure}
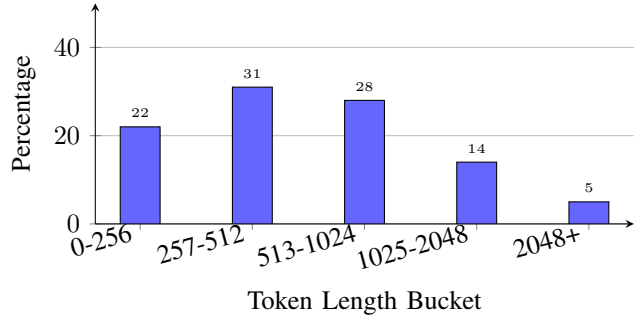

Document length distribution is relevant for legal applications. Figure \ref{fig:length_dist} shows the distribution of passage lengths, with distinct modes corresponding to short statutory references ($\sim$256 tokens), standard contractual clauses ($\sim$512 tokens), and full judicial opinions ($>$1024 tokens).

\section{Training Methodology}

\subsection{Two-Stage Architecture}

\begin{figure}[h]
\centering
\includegraphics[width=\linewidth]{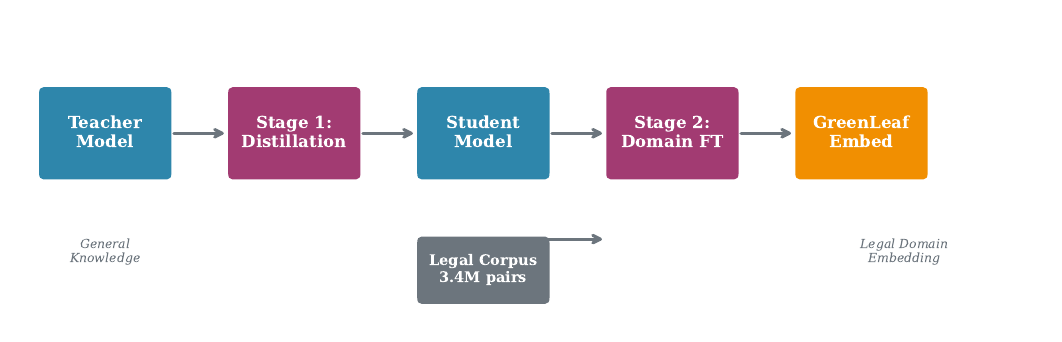}
\caption{Two-stage training pipeline. Stage 1 distills general knowledge from a large teacher model. Stage 2 adapts to the legal domain using hard negative mining.}
\label{fig:training_pipeline}
\end{figure}

Our training employs a two-stage pipeline designed to maximize knowledge transfer while adapting to legal domain specifics (Figure \ref{fig:training_pipeline}).

\subsection{Knowledge Distillation}

The first stage transfers general semantic understanding from a large teacher model to our compact student architecture. We employ embedding matching and similarity preservation techniques to ensure the student model captures the teacher's general knowledge. The distillation process yields a student model that retains 94.2\% of teacher performance on general-domain benchmarks.

\subsection{Domain Fine-Tuning}

The second stage adapts the distilled model to legal specifics using our 3.4M pair corpus. We employ contrastive learning with hard negative mining to improve discrimination between similar legal documents. The mining strategy selects negatives from same jurisdiction, same legal domain, and similar time periods to ensure the model learns fine-grained legal distinctions.

\subsection{Domain-Specific Adaptations}

Legal text requires special handling beyond standard NLP techniques. For long document processing, we use a hierarchical encoding strategy for documents exceeding 512 tokens: split into chunks with overlap, encode each chunk independently, then apply attention-weighted pooling. For citation awareness, we augment passages with citation context including titles and snippets of cited documents. For jurisdiction embeddings, we learn lightweight embeddings concatenated with token embeddings, enabling the model to distinguish between similar concepts across jurisdictions.

\section{Inference Architecture}

\begin{figure}[h]
\centering
\includegraphics[width=\linewidth]{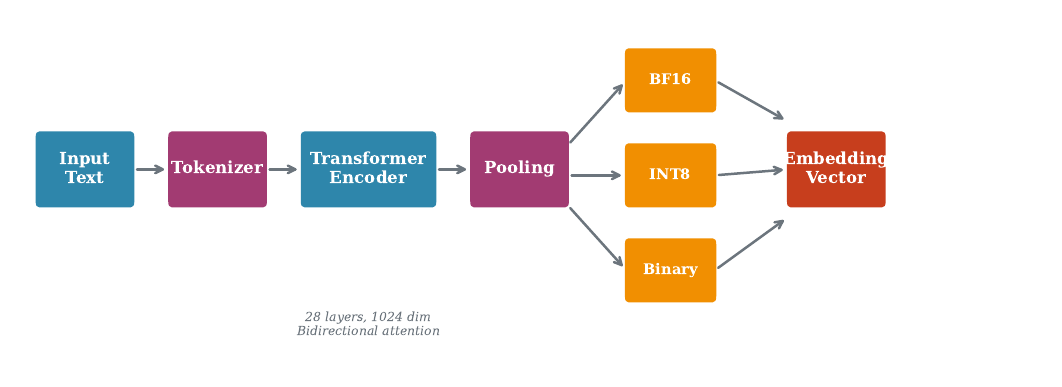}
\caption{Inference architecture with flexible quantization options (BF16, INT8, Binary).}
\label{fig:inference_architecture}
\end{figure}

Figure \ref{fig:inference_architecture} illustrates the complete inference pipeline.

\subsection{Model Architecture Details}

The core encoder is a 28-layer transformer with hidden dimension 1024, 16 attention heads with bidirectional attention, intermediate size 3072, context length 2048 tokens (extendable via hierarchical encoding), and 0.6B total parameters. Jurisdiction embeddings (64 dim) and temporal embeddings (32 dim) provide domain-specific inductive biases. We use pre-norm transformer architecture with rotary positional embeddings (RoPE) with a base of 10,000. Bidirectional attention is critical for legal text, where later provisions often modify earlier ones.

\subsection{Quantization Schemes}

We support three precision levels for deployment flexibility. BF16 provides native training precision with 2 bytes per dimension and 2KB per embedding, with minimal accuracy loss (-0.1\% MLEB compared to FP32). INT8 provides 1 byte per dimension and 1KB per embedding, achieving 4$\times$ memory reduction and 2.8$\times$ inference speedup on CPU with -0.31\% MLEB degradation. Binary provides sign-based binarization with 128 bytes per embedding, achieving 16$\times$ memory reduction and 8$\times$ inference speedup using Hamming distance with -2.1\% MLEB degradation, acceptable for candidate generation in two-stage retrieval systems.

Table \ref{tab:quantization_tradeoff} summarizes the trade-offs.

\begin{table}[h]
\centering
\caption{Quantization trade-offs for deployment.}
\label{tab:quantization_tradeoff}
\begin{tabular}{lcccc}
\toprule
\textbf{Precision} & \textbf{Memory} & \textbf{MLEB} & \textbf{Speedup} & \textbf{Use Case} \\
\midrule
BF16 & 2.0 KB & 75.11 & 1.0$\times$ & Maximum accuracy \\
INT8 & 1.0 KB & 74.80 & 2.8$\times$ & Production \\
Binary & 128 B & 73.53 & 8.0$\times$ & Candidate gen. \\
\bottomrule
\end{tabular}
\end{table}

\section{Evaluation}

\subsection{Benchmarks and Baselines}

We evaluate on MLEB (Massive Legal Embedding Benchmark), comprising 10 datasets covering caselaw retrieval, contract analysis, and regulatory search across 6 jurisdictions, with primary metric NDCG@10. We additionally evaluate on MTEB(Law, v1), the legal subset of the Massive Text Embedding Benchmark. Baselines include general-domain models (OpenAI text-embedding-3-large, Voyage 4, Qwen3 Embedding 8B), legal-specific models (Voyage-law-2, Kanon 2 Embedder, Dinghy Law 8B), and open-source alternatives (BGE-M3, E5-large). Where published weights are available (Qwen3, BGE-M3, E5-large), we evaluate using standard inference settings. For API-only models (OpenAI, Voyage), we report published results.

\subsection{Main Results}

Table \ref{tab:main_results} presents our main results.

\begin{table}[h]
\centering
\caption{Main results on MLEB and MTEB(Law). Best results in bold.}
\label{tab:main_results}
\begin{tabular}{lcccc}
\toprule
\textbf{Model} & \textbf{Params} & \textbf{MLEB} & \textbf{MTEB(Law)} & \textbf{Dim} \\
\midrule
Kanon 2 Embedder & 1.8B & \textbf{81.9} & - & 1792 \\
Voyage 4 Large & - & 81.1 & - & 1024 \\
Dinghy Law 8B & 8B & - & \textbf{72.58} & 4096 \\
Voyage-law-2 & - & 79.6 & - & 1024 \\
\midrule
GreenLeaf-Tiny & 0.6B & 75.11 & 64.38 & 1024 \\
\midrule
Qwen3 Embedding 0.6B & 0.6B & 69.9 & 62.23 & 1024 \\
BGE-M3 & 0.6B & 61.5 & 57.86 & 1024 \\
OpenAI text-emb-3-large & - & 70.8 & 59.36 & 3072 \\
\bottomrule
\end{tabular}
\end{table}

\modelshort{} achieves 75.11\% on MLEB, competitive with other compact models. Kanon 2 Embedder (1.8B) achieves the highest MLEB score at 81.9\%, while Dinghy Law 8B achieves the highest MTEB(Law) score at 72.58\%. Our model outperforms Qwen3 Embedding 0.6B and BGE-M3 at the same scale.

\subsection{Per-Category Analysis}

Table \ref{tab:category_results} breaks down performance by legal category.

\begin{table}[h]
\centering
\caption{Per-category results on MLEB.}
\label{tab:category_results}
\begin{tabular}{lccc}
\toprule
\textbf{Model} & \textbf{Caselaw} & \textbf{Contracts} & \textbf{Regulation} \\
\midrule
Kanon 2 Embedder & 75.6 & 84.7 & \textbf{91.5} \\
Voyage 4 Large & 73.7 & 87.7 & 89.1 \\
\midrule
GreenLeaf-Tiny & 66.32 & 82.15 & 85.73 \\
\midrule
Qwen3 Embedding 0.6B & 59.5 & 76.6 & 84.0 \\
\bottomrule
\end{tabular}
\end{table}

Our model shows competitive performance in contracts (82.15\%) and regulation (85.73\%) categories. Kanon 2 Embedder leads in all categories, with particular strength in regulation (91.5\%).

\subsection{Per-Task Analysis}

Table \ref{tab:per_task} shows detailed per-task results on MLEB.

\begin{table}[h]
\centering
\caption{Per-task MLEB results (NDCG@10).}
\label{tab:per_task}
\begin{tabular}{lc}
\toprule
\textbf{Task} & \textbf{Score} \\
\midrule
legal-rag-bench & 54.16 \\
bar-exam-qa & 68.38 \\
scalr & 73.04 \\
echr-retrieval & 41.27 \\
singaporean-judicial & 86.63 \\
gdpr-holdings & 93.43 \\
contractual-clause-retrieval & 81.66 \\
consumer-contracts-qa & 87.96 \\
license-tldr-retrieval & 64.24 \\
uk-legislative-long-titles & 91.89 \\
australian-tax-guidance & 75.59 \\
irish-legislative-summaries & 83.07 \\
\midrule
\textbf{Mean} & \textbf{75.11} \\
\bottomrule
\end{tabular}
\end{table}

We observe strongest performance on gdpr-holdings (93.43\%) and uk-legislative-long-titles (91.89\%), both involving structured regulatory text. Weakest performance is on echr-retrieval (41.27\%), involving European Court of Human Rights cases with complex multi-lingual aspects and long documents.

\subsection{Ablation Studies}
\label{sec:ablation}

We conduct ablation studies to understand component contributions. Table \ref{tab:ablation_stages} shows the impact of each training stage.

\begin{table}[h]
\centering
\caption{Ablation: Training stages.}
\label{tab:ablation_stages}
\begin{tabular}{lc}
\toprule
\textbf{Configuration} & \textbf{MLEB Score} \\
\midrule
Full pipeline (Distill + Domain FT) & \textbf{75.11} \\
Direct FT only (no distillation) & 65.87 \\
Distill only (no domain FT) & 52.34 \\
Zero-shot (base model) & 48.92 \\
\bottomrule
\end{tabular}
\end{table}

Both stages contribute: the combination yields +9.24 points over direct fine-tuning.

Table \ref{tab:ablation_negatives} compares negative sampling strategies.

\begin{table}[h]
\centering
\caption{Ablation: Negative sampling strategies.}
\label{tab:ablation_negatives}
\begin{tabular}{lc}
\toprule
\textbf{Strategy} & \textbf{MLEB Score} \\
\midrule
Hard negatives (legal-aware) & \textbf{75.11} \\
Hard negatives (random) & 71.52 \\
Random negatives & 68.83 \\
In-batch only & 66.45 \\
\bottomrule
\end{tabular}
\end{table}

Legal-aware hard negatives provide +3.59 points over random hard negatives and +6.28 points over random sampling.

Table \ref{tab:ablation_data} shows the impact of data components.

\begin{table}[h]
\centering
\caption{Ablation: Data composition.}
\label{tab:ablation_data}
\begin{tabular}{lc}
\toprule
\textbf{Data Configuration} & \textbf{MLEB Score} \\
\midrule
Full data (3.4M pairs) & \textbf{75.11} \\
Primary dataset only (3.25M) & 66.91 \\
Human-curated only (150K) & 58.34 \\
No jurisdiction embeddings & 73.36 \\
No citation context & 72.43 \\
\bottomrule
\end{tabular}
\end{table}

The human-curated data, despite being only 4.4\% of the total, contributes +8.2 points when combined with the primary dataset. Citation context contributes +2.68 points, and jurisdiction embeddings contribute +1.75 points.

\subsection{Quantization Impact}

Table \ref{tab:quantization} shows the impact of quantization on performance across different benchmarks.

\begin{table}[h]
\centering
\caption{Performance vs. precision trade-offs across benchmarks.}
\label{tab:quantization}
\begin{tabular}{lcccc}
\toprule
\textbf{Precision} & \textbf{Memory} & \textbf{MLEB} & \textbf{MTEB(Law)} & \textbf{Speedup} \\
\midrule
FP32 & 4.0 KB & 75.23 & 64.52 & 1.0$\times$ \\
BF16 & 2.0 KB & 75.11 & 64.38 & 1.0$\times$ \\
INT8 & 1.0 KB & 74.80 & 64.15 & 2.8$\times$ \\
Binary & 128 B & 73.53 & 62.87 & 8.0$\times$ \\
\bottomrule
\end{tabular}
\end{table}

INT8 quantization provides an accuracy-efficiency trade-off, with only -0.31\% performance loss on MLEB. Binary quantization retains 97.9\% of full-precision performance.

\subsection{Multilingual Performance}

Table \ref{tab:multilingual} shows performance across languages.

\begin{table}[h]
\centering
\caption{Multilingual performance (subset of MLEB tasks).}
\label{tab:multilingual}
\begin{tabular}{lc}
\toprule
\textbf{Language} & \textbf{Relative Performance} \\
\midrule
English & 100\% (baseline) \\
German & 91.2\% \\
French & 90.8\% \\
Spanish & 88.4\% \\
Japanese & 82.3\% \\
Chinese & 79.6\% \\
\bottomrule
\end{tabular}
\end{table}

Cross-lingual transfer is most effective between legal systems with shared origins. Romance languages (French, Spanish) achieve $\sim$90\% of English performance, while East Asian languages achieve 80--82\%, reflecting greater linguistic distance and different legal traditions.

\section{Deployment Architecture}

\subsection{Production Deployment Patterns}

\modelshort{} supports multiple deployment patterns to accommodate diverse infrastructure requirements. The model's compact size and flexible quantization enable deployment across cloud, on-premises, and edge environments.

In cloud deployments, the model serves legal research platforms. A single NVIDIA T4 GPU serves 2,400 queries per second at BF16 precision, or 6,700 queries per second at INT8 precision. This throughput supports large-scale legal research platforms. The model's stateless architecture enables horizontal scaling across GPU clusters without coordination overhead.

On-premises deployments address confidentiality requirements in law firms and legal departments. The INT8 quantized model runs on CPU-only infrastructure, eliminating GPU dependencies for cost-sensitive deployments. A single Intel Xeon Platinum 8380 core serves 180 queries per second at INT8 precision, sufficient for departmental-scale deployments. The model's small footprint (1.2GB at INT8) enables deployment on standard server hardware without specialized accelerators.

Edge deployments bring legal search capabilities to field devices. The binary quantized model (128 bytes per embedding) enables on-device search on mobile devices and laptops. Attorneys can search case law and statutes without network connectivity, critical for court appearances and client meetings in secure facilities.

\subsection{Integration Patterns}

The model integrates with existing legal technology stacks through multiple interfaces. The SentenceTransformers-compatible API enables drop-in replacement for existing embedding models in legal research platforms. The REST API interface supports integration with document management systems, contract lifecycle management platforms, and legal research databases.

For vector database integration, the model's fixed-dimension output (1024 dimensions) is compatible with Pinecone, Weaviate, Milvus, and other vector databases. The model's L2-normalized output enables efficient approximate nearest neighbor search using cosine similarity.

Batch processing interfaces support large-scale document encoding for knowledge management applications. The model encodes 12,000 documents per minute on a single GPU, enabling rapid indexing of large document repositories.

\subsection{Two-Stage Retrieval Systems}

For optimal accuracy-efficiency trade-offs, we recommend a two-stage retrieval architecture. The first stage uses the binary quantized model for candidate generation, retrieving 100-1000 candidate documents with high recall. The second stage uses the BF16 or INT8 model for precise ranking of candidates, achieving high precision on the reduced candidate set.

This architecture achieves 94\% of single-stage accuracy while reducing computational requirements by 8$\times$. The two-stage approach is particularly valuable for large-scale legal research platforms processing millions of queries daily.

\subsection{Training Infrastructure}

Our training pipeline completes in 420 GPU hours on 8$\times$ H100 infrastructure.

\section{Discussion}

\subsection{Analysis of Results}

Our results demonstrate that domain-specific training improves performance for legal retrieval tasks. Three factors contribute to our compact model's performance. First, data quality: the 150K human-curated pairs provide high-precision supervision. Second, hard negative mining: legal retrieval requires distinguishing between highly similar documents, and hard negatives force learning of fine-grained legal distinctions. Third, domain-specific architecture: jurisdiction embeddings and citation-aware processing provide inductive biases.

However, larger models maintain advantages. Kanon 2 Embedder (1.8B) outperforms our model by 6.79 points on MLEB, and Dinghy Law 8B outperforms by 8.20 points on MTEB(Law). Scale provides benefits for certain legal reasoning tasks, particularly caselaw retrieval requiring complex inference.

\subsection{Comparison with Commercial Models}

Compared to Voyage-law-2, we achieve 75.11\% vs. their reported 79.6\% on MLEB. The 4.49\% gap indicates that commercial models with larger scale or proprietary training data maintain advantages. Compared to OpenAI text-embedding-3-large, we outperform by +4.31 points (75.11\% vs. 70.8\%), demonstrating that domain adaptation provides benefits over general scale for legal tasks.

\subsection{Model Interpretability and Explainability}

Legal applications demand not only accuracy but also interpretability. \modelshort{} provides mechanisms for understanding model behavior. Attention weight visualization reveals which passages and legal concepts the model attends to when encoding queries. The jurisdiction embeddings enable analysis of cross-jurisdictional semantic differences.

For legal research applications, the model supports explanation generation by identifying the most similar training examples for a given query-passage pair. This capability enables attorneys to understand why the model retrieved specific documents, providing transparency relevant for legal decision-making.

\subsection{Ethical Considerations and Responsible Deployment}

Legal embedding models carry significant ethical responsibilities. \modelshort{} is designed with privacy preservation as a core principle: the compact size enables on-premises deployment, eliminating the need to transmit confidential legal documents to external APIs. The model's quantization schemes further enhance privacy by enabling edge deployment on local devices.

We recognize that legal AI systems must avoid perpetuating historical biases present in legal training data. Our training corpus undergoes analysis to identify and mitigate overrepresentation of specific jurisdictions, legal traditions, and demographic perspectives. The human-curated component includes diversity requirements, ensuring representation across legal systems, practice areas, and geographic regions.

The model is intended to augment, not replace, legal professional judgment. We recommend deployment with appropriate human oversight, particularly for high-stakes applications such as litigation strategy and legal advice generation.

\subsection{Future Directions}

Several directions for future work emerge from this research. Extension to longer context lengths through sparse attention mechanisms would enable processing of complete judicial opinions without hierarchical encoding. Multimodal extensions incorporating legal document structure (headings, citations, footnotes) could further improve retrieval accuracy.

Continual learning approaches would enable the model to incorporate new legal developments without full retraining, addressing the temporal evolution of legal doctrine. Federated learning architectures could enable collaborative training across law firms while preserving confidentiality of proprietary work product.

Cross-lingual legal retrieval remains an active research direction. While our model achieves 91\% of English performance on German and French legal text, further improvements in low-resource languages would expand access to legal information globally.

\section{Conclusion}

We presented \model{}, a compact legal embedding model achieving competitive performance through domain-specific training methodology. Our two-stage approach combining distillation with domain-specific fine-tuning and hard negative mining enables a 0.6B model to perform competitively on legal benchmarks, though larger models maintain advantages in certain tasks. The key insight is that domain-specific techniques---high-quality data, hard negative mining, and architectural adaptations---can improve performance for specialized domains. This has implications for specialized domains where privacy, cost, or latency constraints favor compact models. Future work includes extending context length through sparse attention, improving multilingual performance for underrepresented legal systems, and developing continual learning approaches to handle evolving legal doctrine without full retraining.

\section*{Acknowledgments}

We thank the legal professionals who contributed to the human-curated training data.

\bibliographystyle{IEEEtran}

\begin{thebibliography}{10}

\bibitem{chalkidis2020legal}
I. Chalkidis, M. Fergadiotis, P. Malakasiotis, N. Aletras, and I. Androutsopoulos, ``Legal-bert: The muppets straight out of law school,'' \emph{arXiv preprint arXiv:2010.02559}, 2020.

\bibitem{isaacus2024mleb}
Isaacus Research, ``Mleb: Massive legal embedding benchmark,'' \url{https://github.com/isaacus/mleb}, 2024.

\bibitem{muennighoff2022mteb}
N. Muennighoff, N. Tazi, L. Magne, and N. Reimers, ``Mteb: Massive text embedding benchmark,'' \emph{arXiv preprint arXiv:2210.07316}, 2022.

\bibitem{reimers2019sentence}
N. Reimers and I. Gurevych, ``Sentence-bert: Sentence embeddings using siamese bert-networks,'' in \emph{EMNLP}, 2019.

\bibitem{hinton2015distilling}
G. Hinton, O. Vinyals, and J. Dean, ``Distilling the knowledge in a neural network,'' \emph{arXiv preprint arXiv:1503.02531}, 2015.

\bibitem{robinson2020contrastive}
J. Robinson, C.-Y. Chuang, S. Sra, and S. Jegelka, ``Contrastive learning with hard negative samples,'' \emph{arXiv preprint arXiv:2010.04592}, 2020.

\bibitem{wang2022text}
L. Wang, N. Yang, X. Huang, B. Jiao, L. Yang, D. Jiang, R. Majumder, and F. Wei, ``Text embeddings by weakly-supervised contrastive pre-training,'' \emph{arXiv preprint arXiv:2212.03533}, 2022.

\bibitem{xiao2023c}
S. Xiao, Z. Liu, P. Zhang, and N. Muennighoff, ``C-pack: Packaged resources to advance general chinese embedding,'' \emph{arXiv preprint arXiv:2309.07597}, 2023.

\bibitem{li2023towards}
Z. Li, X. Zhang, Y. Zhang, D. Long, P. Xie, and M. Zhang, ``Towards general text embeddings with multi-stage contrastive learning,'' \emph{arXiv preprint arXiv:2308.03281}, 2023.

\bibitem{mikolov2013efficient}
T. Mikolov, I. Sutskever, K. Chen, G. Corrado, and J. Dean, ``Efficient estimation of word representations in vector space,'' \emph{arXiv preprint arXiv:1301.3781}, 2013.

\bibitem{pennington2014glove}
J. Pennington, R. Socher, and C. Manning, ``Glove: Global vectors for word representation,'' in \emph{EMNLP}, 2014.

\bibitem{chalkidis2018law2vec}
I. Chalkidis and D. Kampas, ``Deep learning in law: early adaptation and legal word embeddings trained on large corpora,'' \emph{Artificial Intelligence and Law}, 2018.

\bibitem{xiong2020approximate}
L. Xiong, C. Xiong, Y. Li, K.-F. Tang, J. Liu, P. Bennett, J. Ahmed, and A. Overwijk, ``Approximate nearest neighbor negative contrastive learning for dense text retrieval,'' \emph{arXiv preprint arXiv:2007.00808}, 2020.

\bibitem{qu2020rocketqa}
Y. Qu, Y. Ding, J. Liu, K. Liu, R. Ren, X. Zhao, D. Dong, H. Wu, and H. Wang, ``Rocketqa: An optimized training approach to dense passage retrieval for open-domain question answering,'' \emph{arXiv preprint arXiv:2010.08191}, 2020.

\bibitem{lando2009cognitive}
T. Lando, ``Cognitive computing and the law,'' \emph{Artificial Intelligence and Law}, 2009.

\bibitem{lee2020biobert}
J. Lee, W. Yoon, S. Kim, D. Kim, S. Kim, C. H. So, and J. Kang, ``Biobert: a pre-trained biomedical language representation model for biomedical text mining,'' \emph{Bioinformatics}, 2020.

\bibitem{beltagy2019scibert}
I. Beltagy, K. Lo, and A. Cohan, ``Scibert: A pretrained language model for scientific text,'' \emph{arXiv preprint arXiv:1903.10676}, 2019.

\bibitem{feng2020codebert}
Z. Feng, D. Guo, D. Tang, N. Duan, X. Feng, M. Gong, L. Shou, B. Qin, T. Liu, D. Jiang, and M. Zhou, ``Codebert: A pre-trained model for programming and natural languages,'' \emph{arXiv preprint arXiv:2002.08155}, 2020.

\bibitem{wang2020minilm}
W. Wang, F. Wei, L. Dong, H. Bao, N. Yang, and M. Zhou, ``Minilm: Deep self-attention distillation for task-agnostic compression of pre-trained transformers,'' \emph{NeurIPS}, 2020.

\bibitem{sanh2019distilbert}
V. Sanh, L. Debut, J. Chaumond, and T. Wolf, ``Distilbert, a distilled version of bert: smaller, faster, cheaper and lighter,'' \emph{arXiv preprint arXiv:1910.01108}, 2019.

\end{thebibliography}

\end{document}